\documentclass[runningheads]{llncs}

\usepackage{eccv}

\usepackage{eccvabbrv}

\usepackage{graphicx}
\usepackage{booktabs}
\usepackage{multirow}
\usepackage{algorithm}
\usepackage{algorithmic}

\usepackage[accsupp]{axessibility}  % Improves PDF readability for those with disabilities.

\usepackage[pagebackref,breaklinks,colorlinks,citecolor=eccvblue]{hyperref}
\usepackage{orcidlink}

\usepackage{amsmath} % for mathematical symbols and formulas
\usepackage{amssymb} % for additional math symbols
\usepackage{amsfonts} % for math fonts

\begin{document}

\title{MAPPO-PIS: A Multi-Agent Proximal Policy Optimization Method with Prior Intent Sharing for CAVs' Cooperative Decision-Making} 

\titlerunning{MAPPO-PIS}

% 设置脚注样式
\renewcommand{\thefootnote}{\fnsymbol{footnote}}


\author{Yicheng Guo $^1$\thanks{Equal contribution.}
\and
Jiaqi Liu $^{1 \star}$ \and
Rongjie Yu $^{1}$ \and
Peng Hang $^{1}$ \thanks{Corresponding author: \texttt{hangpeng@tongji.edu.cn}}
\and
Jian Sun $^{1}$\\
}

\authorrunning{Y. Guo et al.}

\institute{1 Department of Traffic Engineering and Key Laboratory of Road and Traffic Engineering, Ministry
of Education, Tongji University, Shanghai 201804, China\\
\email{\{guo\_yicheng, liujiaqi13, yurongjie, hangpeng, sunjian\}@tongji.edu.cn}
}

\maketitle

\begin{abstract}
  Vehicle-to-Vehicle (V2V) technologies have great potential for enhancing traffic flow efficiency and safety. 
  However, cooperative decision-making in multi-agent systems, particularly in complex human-machine mixed merging areas, remains challenging for connected and autonomous vehicles (CAVs). 
  Intent sharing, a key aspect of human coordination, may offer an effective solution to these decision-making problems, but its application in CAVs is under-explored. 
  This paper presents an intent-sharing-based cooperative method, the Multi-Agent Proximal Policy Optimization with Prior Intent Sharing (MAPPO-PIS), which models the CAV cooperative decision-making problem as a Multi-Agent Reinforcement Learning (MARL) problem.  It involves training and updating the agents' policies through the integration of two key modules: the Intention Generator Module (IGM) and the Safety Enhanced Module (SEM).
  The IGM is specifically crafted to generate and disseminate CAVs' intended trajectories spanning multiple future time-steps. On the other hand, the SEM serves a crucial role in assessing the safety of the decisions made and rectifying them if necessary.
  Merging area with human-machine mixed traffic flow is selected to validate our method.
  Results show that MAPPO-PIS significantly improves decision-making performance in multi-agent systems, surpassing state-of-the-art baselines in safety, efficiency, and overall traffic system performance.
  The code and video demo can be found at: \url{https://github.com/CCCC1dhcgd/A-MAPPO-PIS}.
  
\end{abstract}

\section{Introduction}
\label{sec:intro}

Recent advancements in Connected and Autonomous Vehicle (CAV) technologies have revolutionized the transportation industry by enabling vehicles to communicate and collaborate, thereby enhancing overall traffic management\cite{44,45,liu2024delay}.
One of the critical aspects of this revolution is cooperative decision-making.
With Vehicle-to-Vehicle (V2V) communication, CAVs can improve their perception and safety capabilities by sharing sensor information, thus making efficient decisions in dynamic environments, improving safety and traffic flow\cite{43}.
However, despite the great potential of cooperative decision-making technologies, current methods still face great challenges in the real world, 
like exploring what information to share (\eg vehicle position, signal phase and planned vehicle trajectory), how to share and how to utilize those information to induce the optimal cooperative driving behavior.
One promising approach is intent sharing which proposed by Cooperative Driving Automation(CDA) Committee\cite{13}.
As it in human society, where drivers use signal light to express their intention, namely their future behavior, CAVs can transform and coordinate their intent using V2V technologies.
By sharing CAVs' intention directly, they can achieve better coordination thus avoid conflicts. 
Most previous models, however, focus on inferring other agents intentions, rather than sharing.
In that case, existing models may induce sub-optimal coordination and performance as mistakes can be made when inferring.

\begin{figure}[tb]
  \centering
  % \graphicspath{ {pictures/} }
  \includegraphics[scale = 0.8]{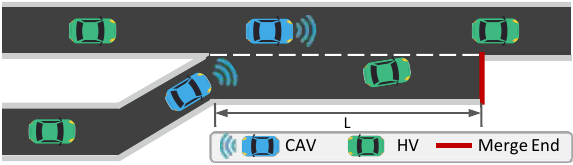}
  \caption{Illustration of merging area with human-machine mixed traffic flow.
  }
  \label{fig:ramp}
\end{figure}

Merging areas, where have complex traffic interactions between vehicles on the ramp or on the through lanes, are one of the most critical recurring bottlenecks\cite{54}. 
In merging area, vehicles from an on-ramp must smoothly integrate into the through lane traffic flow (which is shown in \cref{fig:ramp}), requiring effective collaboration among them.
With the increasing density of vehicles, the merging area bottleneck tends to breakdown, which is a transition from free traffic flow to congested traffic, and then lead to serious traffic delays and accident rates increase\cite{55}.
It is particularly challenging for vehicles to make decision in the merging area, especially in the complex human-machine mixed merging area, thus making it an ideal scenario to test cooperative decision-making method we proposed.

To cope with challenges mentioned above, we propose a novel and efficient method, Multi-Agent Proximal Policy Optimization Method with Prior Intent Sharing for CAVs' Cooperative Decision-Making (\textbf{MAPPO-PIS}), which formulates the decision-making problem of CAVs in human-machine mixed scenario as a decentralized multi-agent RL problem. 
In MAPPO-PIS, each CAV in merging area is modeled as an agent, which has ability to communicate with each other, share important information (like position, speed and intention), and cooperate with other agents. 
We first adopt Multi-Agent Proximal Policy Optimization (MAPPO) as the baseline algorithm, where all agents follow a centralized training and distributed execution (CTDE) strategy framework. 
Subsequently, an \textbf{I}ntention \textbf{G}enerator \textbf{M}odule (\textbf{IGM}) is designed, which could generate and share the intent trajectories of CAVs for multiple time-steps in the future.
Ultimately, we introduce a \textbf{S}afety \textbf{E}nhanced \textbf{M}odule (\textbf{SEM}) to anticipate and identify potential safety risks posed by neighboring vehicles during CAV exploration. This module corrects any detected hazards, thereby ensuring safety and enhancing the algorithm's learning efficiency.
The motivation of these two modules is to enhance the safety performance of CAVs in merging area, using intent sharing. 

In this work, the complex human-machine mixed merging area is selected as the benchmark scenario to validate our proposed method. In addition to experiments under different traffic densities, we also conduct them with heterogeneous vehicles.
Numerical result shows that the proposed MAPPO-PIS significantly outperforms other
existing MARL methods in terms of safety, efficiency, and overall performance of the traffic system.

The contributions of this paper are summarized as follows:
\begin{enumerate}
    \item A novel and efficient algorithm, MAPPO-PIS
    % \footnote{MAPPO-PIS is open sourced in \url{https://github.com/CCCC1dhcgd/A-MAPPO-PIS}} 
    is proposed, where each CAV is modeled as the RL agent and shares driving intention by Intention Generator Module (IGM) to facilitate cooperative decision-making.
    \item The Safety Enhanced Module (SEM) is designed to identify potential safety risks and correct any detected hazards, thereby ensuring safety and enhancing the algorithm's learning efficiency.
    \item The extensive experiments compared with different baselines are conducted, demonstrating that our approach effectively delays and alleviates bottleneck effects while accelerating learning efficiency and enhancing driving safety.
\end{enumerate}

\section{Related work} \label{Related work}

\subsection{CAVs Cooperation}
Cooperative perception, cooperative decision-making and cooperative control are three main components of Vehicle-to Everything (V2X) communication system and in this paper, we mainly focus on how to achieve cooperative decision-making efficiently and safely.
Numerous methods have been explored to address vehicle coordination, which are summarized as follows:

\textbf{Rule-based methods.}
Heuristic rules are always employed in rule-based methods.
Dresner and Stone proposed a reservation scheme to control vehicles traveling on a single intersection of two roads\cite{46}.
Ntousakis\etal proposed a decentralized algorithm for automatic merging control, based on a "first come, first serve" basis to decide the merging sequence\cite{47}.
While these approaches are simple and logical, they are inefficient as traffic demands increase.

\textbf{Optimization-based methods.}
In\cite{48}, the interaction of vehicles are represented as a dynamic system, with actions of controlled vehicle serving as inputs,
and a model predictive control (MPC) method is also proposed to control autonomous vehicles in the merging area.
Although the result of these methods are promising, they fail to meet the requirement of real-time application as accurate computations are needed.

\textbf{Learning-based models.}
Methods like  Neural Network(NN)\cite{49,50} and Reinforcement Learning(RL)\cite{23,28,51} have been explored for cooperative decision.
Although these approaches are effective in simulating interaction dynamics and have effective reasoning capabilities, they are poor in interpretability and generalization ability\cite{5}.

\subsection{Intent and intent sharing}
The intent communication problem has been gaining more attention in the past few years with the proposal of Intent-Sharing Cooperation concept in SAE J3216\cite{13}. 
Past research mostly focused on communication between vehicle and pedestrian\cite{14,15,16}, or between vehicles\cite{17,18,19,20,21,22}. 
In this section, we will mainly discuss about literature on communication between vehicles, namely intent sharing.

\textbf{Intent inferring.} Most of previous papers focused on inferring the intent of other agents\cite{17,18,19,20}. In\cite{17}, Qi and Zhu use the target locations of agents as their intents, which are inferred based on their observation history. In\cite{20}, Wu \etal proposed iPLAN, allowing agents to infer nearby drivers' intents solely from their local observations.
Intent inferring utilizes agents' observation history or traits history to predict their goal.
Intents, however, are future-oriented information that is not revealed by previous actions.

\textbf{Intent sharing.} Researchers adopt various ways to achieve explicit communication\cite{21,22}.
In\cite{22}, Mahajan and Zhang solve the intent-sharing problem in the setting of two-AV, one-way intent-sharing between an intent-sender AV and an intent-receiver AV, and intention of intent-sender AV is described as a set of feasible actions during entire duration of a simulation episode, which must be complied. However, the simulation environments' simplification of human drivers' behaviors and the small number of CAVs may lead to performance gaps between simulation and real-world scenarios.

\subsection{Multi-Agent Reinforcement Learning (MARL)}
Multi-Agent Reinforcement Learning (MARL) is an emerging research field focusing on systems with multiple interacting agents (\eg robots, machines, cars \etc) within a shared environment\cite{12}. MARL algorithms have shown effectiveness in various multi-agent systems, solving real-world problems like traffic control\cite{8}, autonomous driving decision-making\cite{4,6,9}, gaming\cite{10}, and resource allocation\cite{11}.

Early methods in MARL, such as Independent Learning (IQL)\cite{23}, assumed that agents learn independently and treat others as part of the environment.
Although simple and fully scalable, these methods suffer from non-stationarity and partial observability issues\cite{4}. % In\cite{27}, Qmix,  one of Q-learning extensions, decompose the global Q-values into individual agent Q-values through a mixed network, improving scalability and cooperation among agents. However, it faces the limitations in dynamic environments.  
In \cite{26}, parameter sharing strategies have been adopted, extending single-agent RL methods like Proximal Policy Optimization(PPO)\cite{24} and 
Actor Critic using Kronecker-factored Trust Region (ACKTR)\cite{25} to multi-agent setting (\ie Multi-Agent Proximal Policy Optimization (MAPPO)\cite{53} and Multi-agent Actor Critic using Kronecker-factored Trust Region (MAACKTR)) and the experiments show that both algorithms have excellent performance. However, these algorithms still fall short in ensuring adequate safety and reliability, limiting their application in complex scenarios.

To address those challenges and achieve safe and efficient cooperation of vehicles, in this work, the MARL framework is improved by incorporating intention-sharing with priority and proposing a safety-enhanced module.

\section{Methodology}
\begin{figure}[tb]
  \centering
  % \graphicspath{ {pictures/} }
  \includegraphics[scale = 0.5]{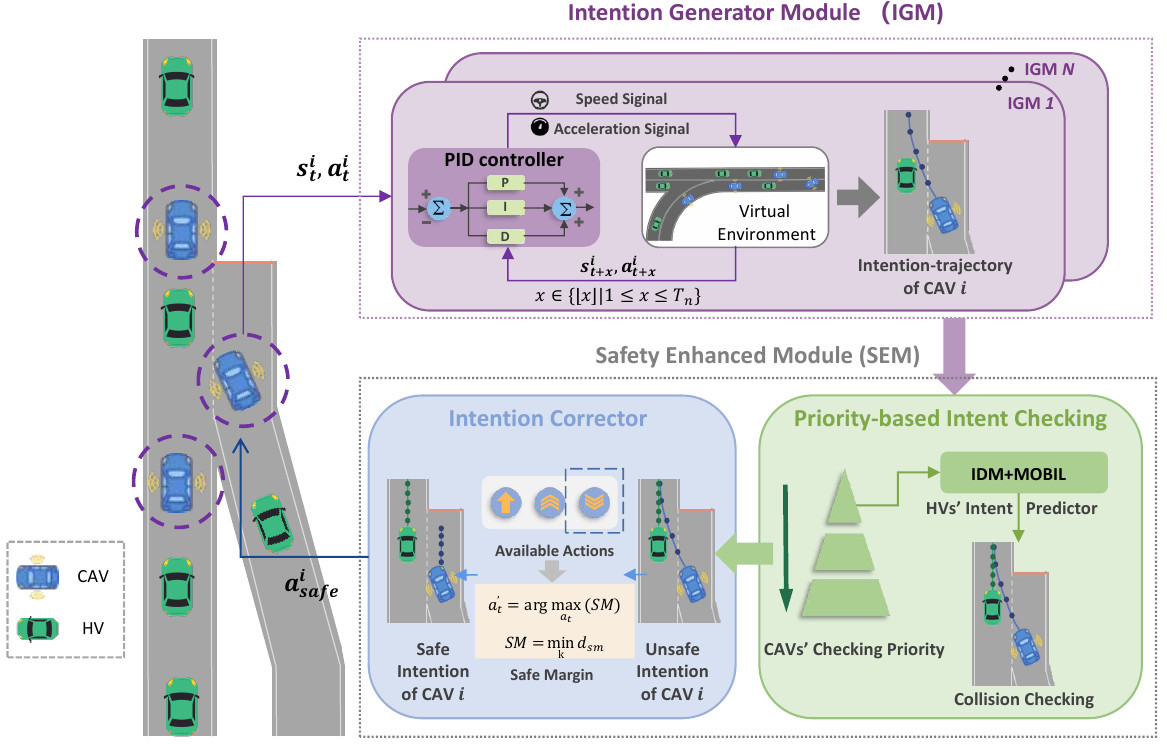}
  \caption{The overview of MAPPO-PIS architecture
  }
  \label{fig:over_frame}
\end{figure}

In this section, the MARL problem is first formulated. Then the MAPPO-PIS method is introduced, which integrates the Intention Generator Module (IGM) and Safety Enhanced Module (SEM). The framework overview is shown in \cref{fig:over_frame}.

\subsection{Problem Formulation} \label{3.1 Problem Formulation}
In this paper, we consider a partially observable Markov decision process (POMDP)\cite{29},
which is consistent with the reality that one vehicle can only sense and cooperate with surrounding vehicles,
and assume that communication is available among all the $N$ agents.
At time step $t$, Agent $i$ observes its state ${s}_{i}$ and select an action ${a}_{i}\in {\mathcal{A}}_{i}$. 
After that, Agent $i$ generate its own intention ${\tau }_{i}$, which could be shared to agents around it.
The joint actions of agents ${a}_{t}=({a}_{t}^{1},\cdots ,{a}_{t}^{N})$ yield the new environment state ${s}'$ and immediate reward ${\left \{{r}_{t}^{i} \right \}}_{i=1}^{N}$ 
according to the transition probability $\mathcal{T}:\mathcal{S}\times\mathcal{A}\times\mathcal{S}\rightarrow [0,1]$
and reward function $\mathcal{S}\times\mathcal{A}\to \mathbb{R}$, respectively,
in which $\mathcal{S}$ and $\mathcal{A}$ are the state space and action space of system (\ie $\mathcal{S}=\textstyle\prod_{i=1}^{N}{\mathcal{S}}_{i}$ and $\mathcal{A}=\textstyle\prod_{i=1}^{N}{\mathcal{A}}_{i}$).
The ultimate goal of each agent is to learn an optimal policy ${\pi }^{*}$ which can maximize the expected return
\begin{equation}
    {R}_{i}=\textstyle\sum_{t=0}^{T}{\gamma }^{t}{r}_{i}({s}_{t},{a}_{t}^{i}).
\end{equation}
where $\gamma$ denotes the discount factor.
Due to page limitations, the detailed explanation of problem formulation is put in \cref{Problem Formulation}.

\subsection{Intention Generator Module (IGM)} \label{IGM}
As described in Section \ref{Related work}, intent is defined as the future behavior of CAVs. 
Therefore, in the intention generator module (IGM), each agent can produce its intention in a time horizon ${T}_{n}$ in the form of its future traveling trajectories based on the next moment action. For agent $i$ at time $t$, we define its intention as:
\begin{equation}
  {\tau }_{i}(t)=\left \{ {p}_{i}(t),{v}_{i}(t),{\theta }_{i}(t)\right \}.
  \label{eq:intention trajectory}
\end{equation}
where ${p}_{i}(t)=({x}_{i}(t),{y}_{i}(t))$ represents the position of the agent $i$, ${v}_{i}(t)$ denotes the velocity of agent $i$, and the ${\theta }_{i}(t)$ denotes the heading of it.

As shown in \cref{fig:over_frame}, firstly, an agent observe the state of the current traffic scenario, and then, high-level decision of agent is made by the MARL agent with the actions defined in Section \ref{Action}. 
Secondly, by inputting the high-level acceleration and lane-change decisions into the IGM, IGM can generate agent's intention-trajectory by utilizing the low-level PID controllers and Proportional Controllers for several iterations.
Finally, all the agents high-level actions and their intention-trajectories will be send to the SEM, in order to enhance their safety performance.

\subsection{Safety Enhanced Module (SEM)}
To improve the driving safety and efficiency of the ego vehicles in complex traffic scenarios, 
A safety enhanced module (SEM) is designed, 
which can be seen as an internal active conflict adjustment mechanism, including two sub-modules: Priority-based Intent Checking and Intention Corrector.

With SEM, the unsafe and inefficient exploration behavior of agent can be corrected, effectively ensuring the stability and continuity of the algorithm. 
The  pseudo code of SEM (\ie \Cref{alg:SEM algorithm}) is shown in detail in \Cref{Re: PSEUDO CODE}. 
The specific calculation process is explained as follow:

\textbf{Priority-based Intent Checking.}
In order to detect collisions between vehicles efficiently, we set a priority list to CAV based on driving principle,
supervising the intention trajectories of CAVs and their surrounding vehicles in the next ${T}_{n}$ steps.

Firstly,  the formula is designed to calculate the priority score of each CAV (\ie Line 2-10 in \Cref{alg:SEM algorithm}):
\begin{equation}
  {p}_{i}={\alpha }_{1}{p}_{m}+{\alpha }_{2}{p}_{e}+{\alpha }_{3}{p}_{h}+{\sigma }_{i}.
  \label{eq:priority}
\end{equation}
where ${p}_{i}$ denotes the priority of the agent $i$, ${\alpha }_{1}$, ${\alpha }_{2}$ and ${\alpha }_{3}$ are the positive weighting parameters for merging metric ${p}_{m}$, merging-end metric ${p}_{e}$ and time headway metric ${p}_{h}$, respectively.
A random variable ${\sigma }_{i}\sim\mathcal{N}(0,0.001)$ is added as a small noise, thus avoiding the problem of the same priority value.
Specifically, ${p}_{m}$ is defined as:
\begin{equation}\label{Eq:pm}
  {p}_{m}=\left\{\begin{matrix}
0.5,& \text{If on merge lane,}\\
 0,& \text{Otherwise.}
\end{matrix}\right.
\end{equation}
which shows that vehicles on the merge lane should be prioritized over those on the through lane due to urgent merging task.
And then, as the merging vehicles which closer to the end of the merge lane should be given higher priority due to a greater risk of collision and deadlocks, the merge-end priority value is calculated as follow:
\begin{equation}\label{Eq:pe}
  {p}_{e}=\left\{\begin{matrix}
\frac{x}{L},& \text{If on merge lane,}\\
 0,& \text{Otherwise.}
\end{matrix}\right.
\end{equation}
where $L$ is the total length of the merging lane, and $x$ is the position of the ego vehicle which on the ramp.
Finally, we defined the time-headway priority as:
\begin{equation}\label{Eq:ph}
  {p}_{h}=-\log_{}{\frac{{d}_{headway}}{{t}_{h}{v}_{t}}}.
\end{equation}
which is similar to the setting of \cref{Eq:rh}, showing vehicles with smaller time-headway are more dangerous than other vehicles.

With the priority scores of CAVs calculated above, at time step $t$, the SEM first produces a priority list ${P}_{t}$, which consists of a list of priority scores and their corresponding ego vehicles (Line 8-9). 
And then, the SEM will sequentially check the intent of the vehicles in the list (Line 14-15).
More specifically, first of all, the selected ego vehicle will predict its surrounding vehicles (only include HVs, CAVs' intention is already generated in Subsection \ref{IGM}).
Secondly, based on its own intention trajectory and the intention trajectories of all other agents around it, the ego vehicle can examine whether its intention trajectory will conflict with its neighboring vehicles (both CAVs and HVs) in a time horizon ${T}_{n}$.
In this paper, we use Intelligent Driver Model (IDM)\cite{35}to predict the  the longitudinal acceleration of HVs, based on the current speed and distance headway. And utilizing MOBIL lane change model\cite{36} to predict the lateral behavior of HVs.

\textbf{Intention Corrector.}
If there is no collision, the safe intentions generated by agents will be realized through low-level PID controllers, and then all the vehicles trajectories will be propagated based on Kinematic bicycle model\cite{37}.
However, if the intention trajectory of CAV overlaps with that of other vehicle, the intention is considered unsafe, and it will be replaced with a new "safe" intention (Line 16-19), which is defined as follow:
\begin{equation} \label{Eq: action replace}
  {a}_{t}^{'}=\arg\max_{{a}_{t}\in {A}_{available}}({\min_{k\in {T}_{n}}{{d}_{sm,k}}}).
\end{equation}
where ${A}_{available}$ is a set of available actions at time step k for the selected agent, ${d}_{sm,k}$ is the safety margin at prediction time step $k$. 
The safety margin is defined as follow:
\begin{equation}
  {d}_{sm,k}=\left\{\begin{matrix}
\min{\left \{|{P}_{{v}_{t},k}-{P}_{{v}_{e},k}|,|{P}_{{v}_{c},k}-{P}_{{v}_{e},k}| \right \}},& \text{If change lane,}\\
 {P}_{{v}_{pre},k}-{P}_{{v}_{e},k},& \text{Otherwise.}
\end{matrix}\right.
\end{equation}
where ${P}_{{v}_{t},k}$ and ${P}_{{v}_{c},k}$ denote the longitudinal positions of the preceding and following vehicles relative to the ego vehicle, respectively, on both the target and current lanes, ${P}_{{v}_{pre},k}$ represents the position of preceding vehicle at time step $k$, and ${P}_{{v}_{e},k}$ is the position of the ego vehicle.

\subsection{Update and Optimization}
In our work, we implement the proposed IGM and SEM on the top of MAPPO, which is a robust MARL algorithm for diverse cooperative tasks.
The whole process of CAVs cooperation is described in \cref{3.1 Problem Formulation},
and for each CAV, it consists of two models, namely actor and critic, which are parameterized by deep neural networks (DNN).
For CAV $i$, let ${\pi}_{\theta}^{i}$ represent the actor network for approximating the policy, and ${V}_{\phi }^{i}$ denote the critic network to approximate the value function, in which $\theta$ and $\phi$ denote the corresponding parameters of the actor network and the critic network.
The policy network of CAV $i$ is updated using the clip objective, which could ensure the update is within a safe range. The formula is:
\begin{equation}
  {L}_{i}^{CLIP}(\theta )=\hat{\mathbb{E}}_{t}[\min(\frac{{\pi }_{\theta }^{i}({a}_{t}^{i}\mid{s}_{t}^{i} )}{{\pi }_{{\theta }_{old}}^{i}({a}_{t}^{i}\mid{s}_{t}^{i} )}\hat{{A}_{t}^{i}},\text{clip}(\frac{{\pi }_{\theta }^{i}({a}_{t}^{i}\mid{s}_{t} ^{i})}{{\pi }_{{\theta }_{old} }^{i}({a}_{t}^{i}\mid{s}_{t}^{i} )},1-\epsilon ,1+\epsilon )\hat{{A}_{t}^{i}})].
\end{equation}
where $\frac{{\pi }_{\theta }^{i}({a}_{t}^{i}\mid {s}_{t}^{i})}{{\pi }_{{\theta}_{old}}^{i}({a}_{t}^{i}\mid {s}_{t}^{i})}$ is the ration of the new policy to the old one, $\epsilon $ is a clip fraction,and $\hat{{A}_{t}^{i}}$ is the general advantage estimation (GAE), which represents the relative value of an action compared to the expected value.
It is formulated by:
\begin{equation}
  {A}_{t}^{i}=\displaystyle\sum_{t=0}^{\infty}(\gamma \lambda )^l{\delta }_{t+l}^{V}.
\end{equation}
where $\gamma$ is the discount value, $\lambda$ denotes the weight value of GAE, and ${\delta }_{t}^{i} = {r}_{t}^{i}+\gamma V({S}_{t+1})-V({S}_{t})$.

\begin{figure}[tb]
  \centering
  % \graphicspath{ {pictures/} }
  \includegraphics[height= 5cm]{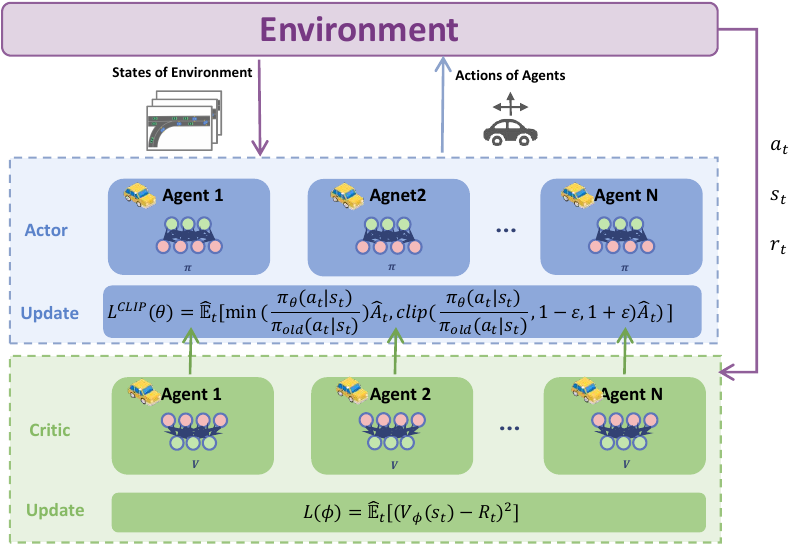}
  \caption{The framework of the Multi-Agent Proximal Policy Optimization (MAPPO).
  }
  \label{fig:Mappo}
\end{figure}

To minimize the value loss, the centralized action-value function ${V}_{\phi}({s}_{t})$ is updated as follows:
\begin{equation}
  L(\phi)=\hat{\mathbb{E}_{t}}[({V}_{\phi}({s}_{t})-{R}_{t})^2].
\end{equation}
where ${R}_{t}$ is the cumulative return and ${V}_{\phi}({s}_{t})$ is the current value estimate. The schematic diagram is shown in figure \ref{fig:Mappo}.

\section{Experiment Results and Analysis}

\subsection{Experimental Setups}
{\bf Simulation Settings.}
To evaluate the performance of MARL algorithm we proposed, the merging scenario in human-machine mixed driving environment is designed, which is illustrated in \cref{fig:ramp}. 
Besides, to make the scenario more realistic, the heterogeneity of vehicles is considered (\ie. Aggressive, Normal and Timid\cite{5}). 
Overall, four levels of traffic scenarios are considered in this work, which are defined as follow:
\begin{itemize}
\item Easy mode: 1-3 CAVs and 1-3 homogeneous HVs (\ie All the HVs' driving styles are the same) are randomly generated on the through lane or ramp, and drive at random speeds.
\item Hard mode: 3-6 CAVs and 3-6 homogeneous HVs.
\item Easy mode with heterogeneous vehicle: 1-3 CAVs and 1-3 heterogeneous HVs. The driving style for each HV is generated randomly.
\item Hard mode with heterogeneous vehicle: 3-6 CAVs and 3-6 heterogeneous HVs. The driving style for each HV is generated randomly.
\end{itemize}

In the training process, we train all the algorithms over 2 millions steps with 3 random seeds (\ie 0, 1000 and 2024), and evaluate them 3 times every 200 training episodes. 
Curriculum learning, which can utilized prior knowledge before training, is employed to faster the training process in hard mode scenarios\cite{4}.
In the simulation environment, the initial speed of all vehicles is set to 25 $m/s$ with random noise (\ie 0~2 $m/s$), and we set the coefficients ${\omega }_{c},{\omega }_{s},{\omega }_{h}$ and ${\omega }_{m}$ as 200, 1, 4 and 4, respectively.
In IGM, the time horizon ${T}_{n}$ is set to 8.
In SEM, the priority parameters of CAVs ${\alpha }_{1}$, ${\alpha }_{2}$ and ${\alpha }_{3}$ are set to 1, 1 and 0.5, respectively.
Here, we utilized  several state-of-the-art MARL algorithms as our baselines, namely MAPPO, MAACKTR and Multi-agent advantage actor-critic (MAA2C)\cite{52}.

The merging environment are simulated by the highway-env simulator\cite{40}. In the experiments, we adopted the default setting of IDM, MOBIL model, and heterogeneous vehicles which can be found in the simulator.

\subsection{Curriculum Learning}
In experiments, curriculum learning is adopted in the hard mode, which could use prior knowledge before training and also extract dynamics information during training\cite{41}, thus speed up the training process and also improve the performance of algorithm.
To be specific, we utilize the trained model in the easy mode and continue to train the model in hard mode, rather than training a model from scratch. 

\begin{figure}[tb]
  \centering
  \begin{subfigure}{0.48\linewidth}
    \includegraphics[scale = 0.21]{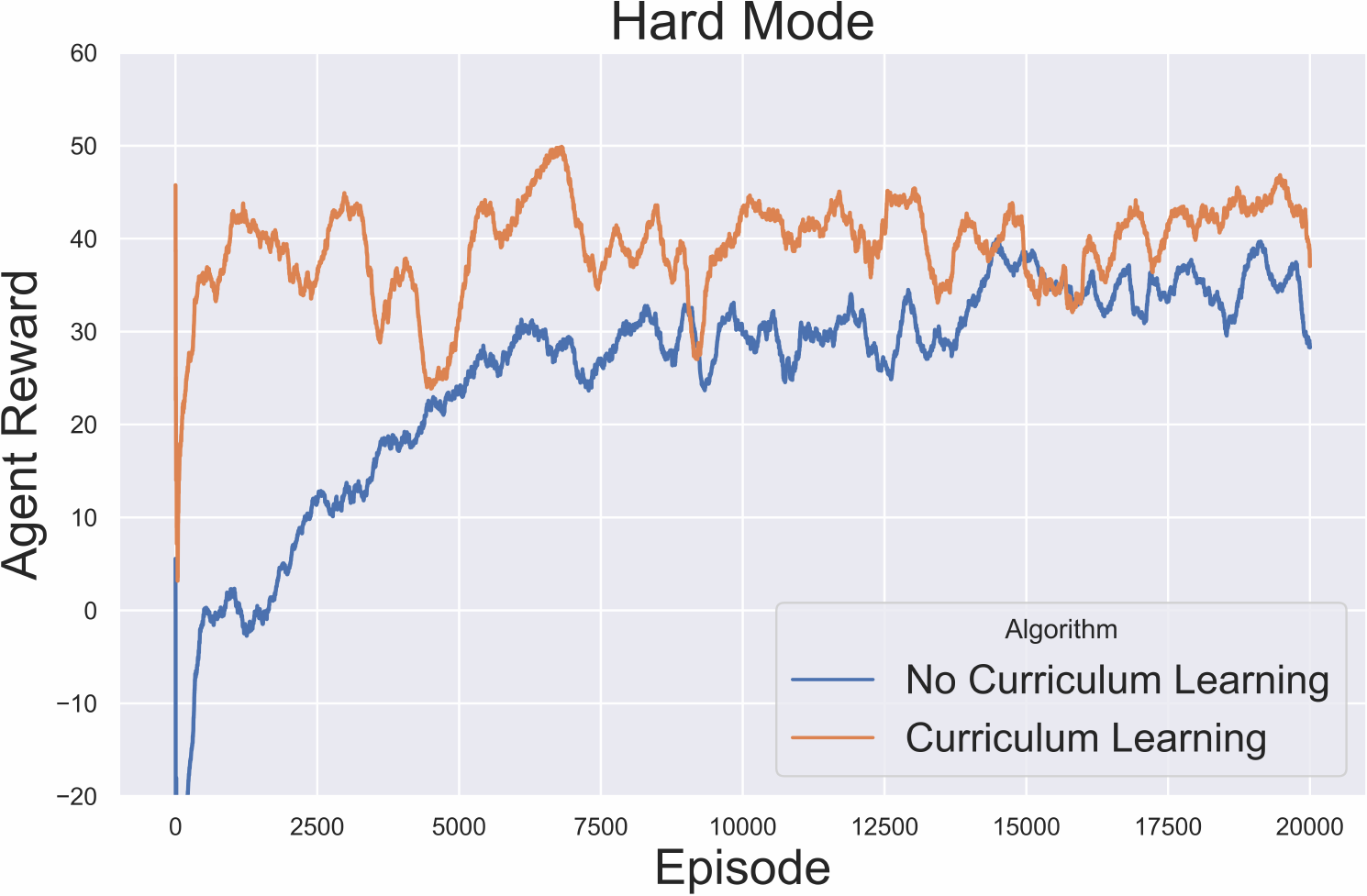}  
    \caption{Training reward curves}
    \label{fig:Curriculum_Learning}
  \end{subfigure}
  \hfill
  \begin{subfigure}{0.48\linewidth}
    \includegraphics[scale = 0.21]{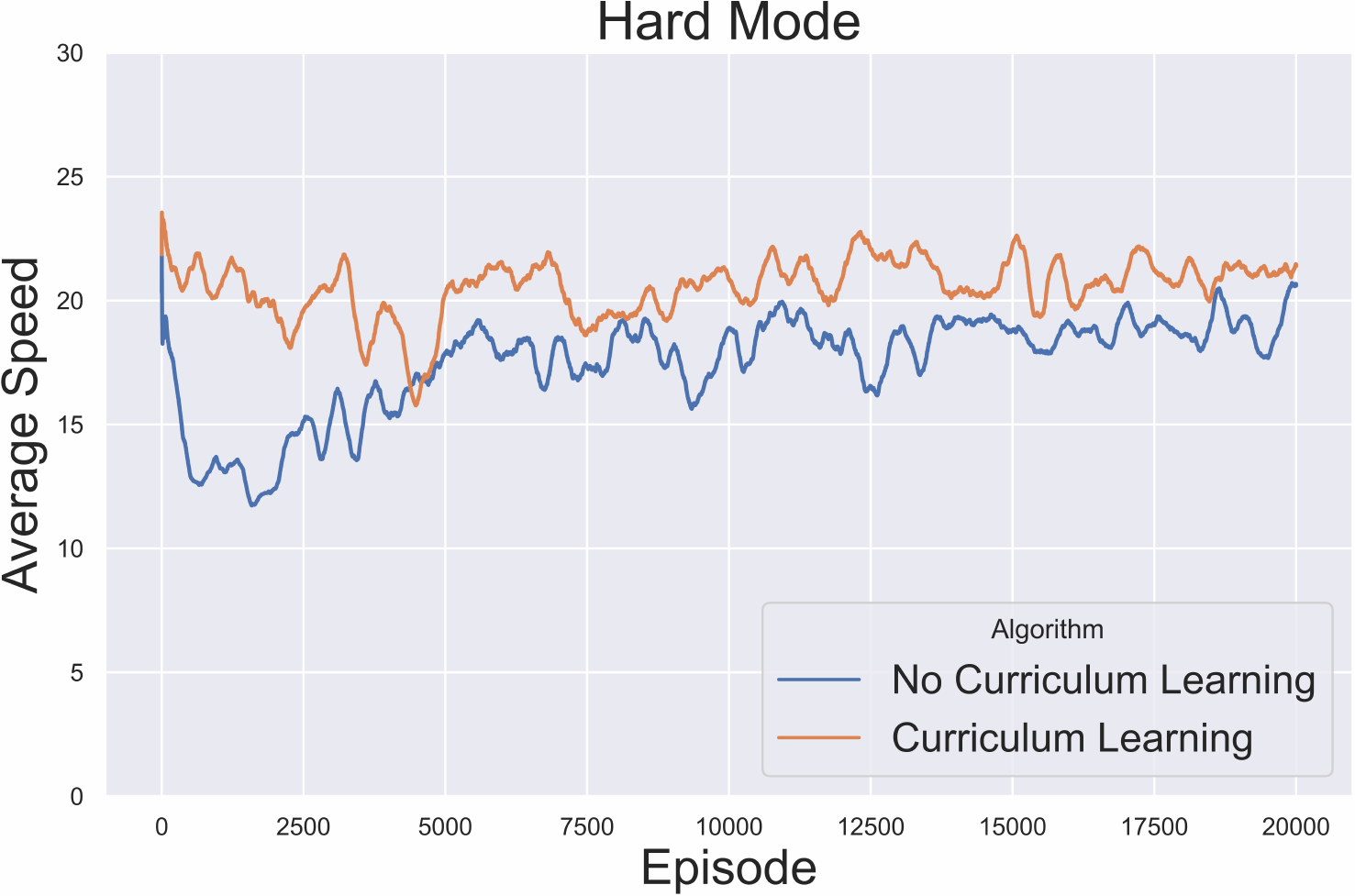}
    \caption{Average speed curves during training}
    \label{fig:Speed_curriculum}
  \end{subfigure}
  \caption{Performance of MAPPO-PIS with and without
curriculum learning for Hard Mode, and the seed is set to 0.}
  \label{fig:Curriculum}
\end{figure}

\cref{fig:Curriculum} shows the training performance comparison between MAPPO-PIS with and without curriculum learning in the hard mode.
In \cref{fig:Curriculum_Learning}, it is clear that curriculum learning could help the algorithm to achieve better performance and faster convergence.
We can also see slight improvement in \cref{fig:Speed_curriculum}, as the algorithm with curriculum learning is on average 4m/s faster than the one without curriculum learning. 

\subsection{Ablation Study}
To evaluate the benefit of our approach, we conduct experiments under two settings, \ie MAPPO with and without IGM and SEM, and evaluate the performance of both algorithm in terms of efficiency, safety and robustness under different scenarios.

\begin{figure}[tb]
  \centering
  \begin{subfigure}{0.48\linewidth}
    \includegraphics[scale = 0.21]{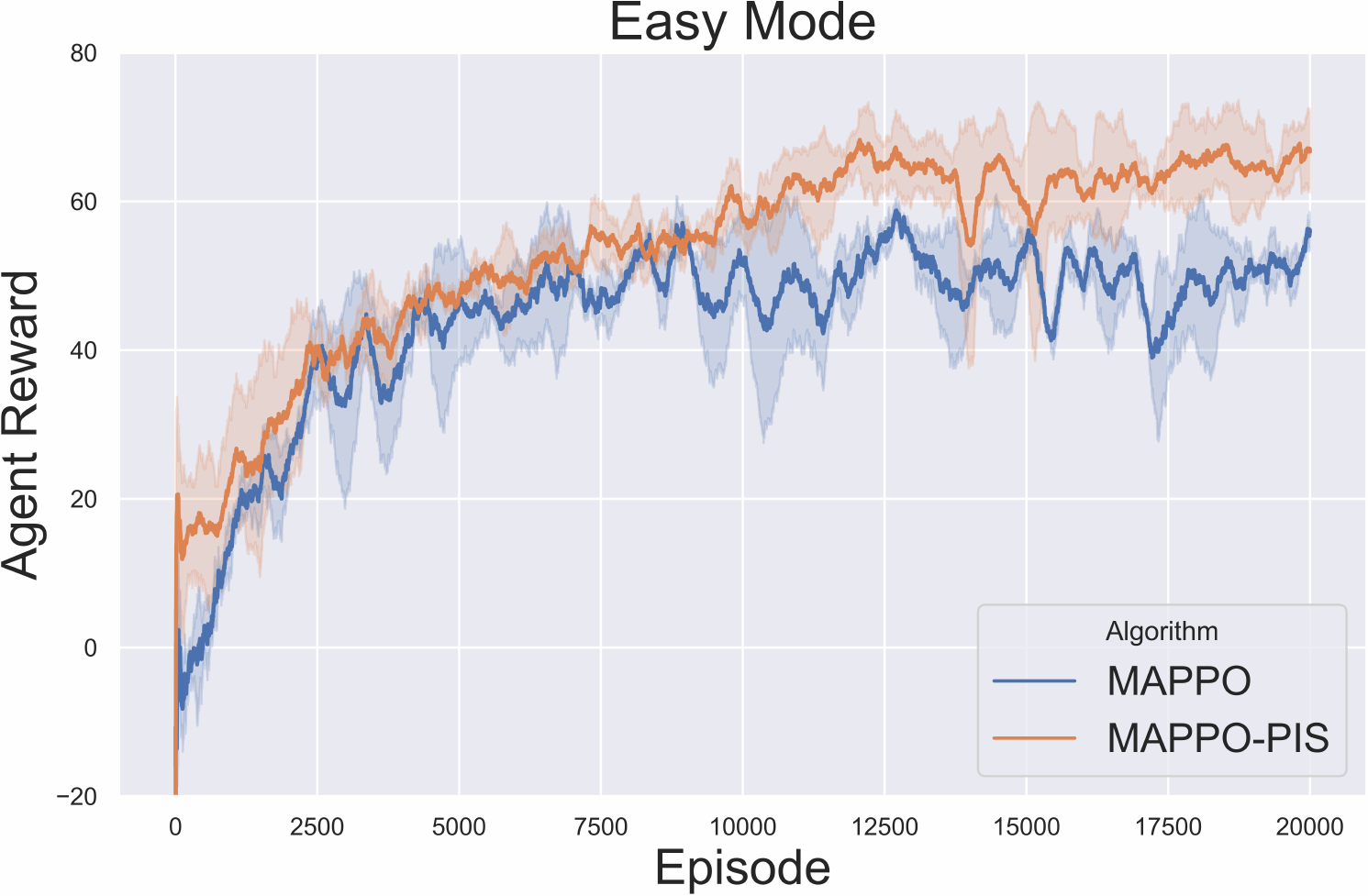} 
    \caption{}
    \label{fig:ab_reward_1}
  \end{subfigure}
  \begin{subfigure}{0.48\linewidth}
    \includegraphics [scale = 0.21]{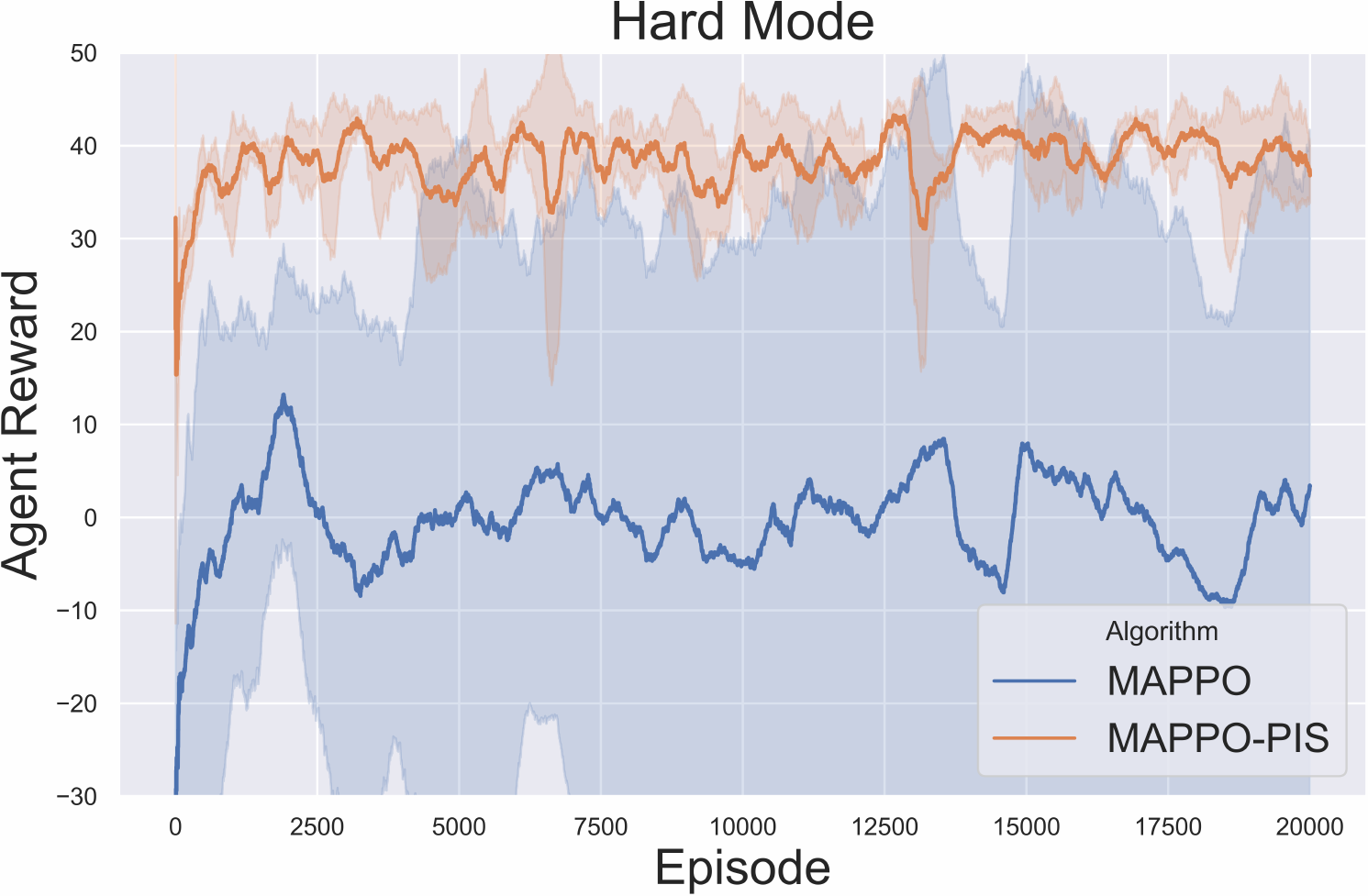}
    \caption{}
    \label{fig:ab_reward_2}
  \end{subfigure}
  \caption{Training average reward curves under different levels of traffic scenario. The shadow region of curves is the confidence interval within the standard deviation.}
  \label{fig:ablation_reward}
\end{figure}

\begin{figure}[tb]
  \centering
  \begin{subfigure}{0.48\linewidth}
    \includegraphics[scale = 0.21]{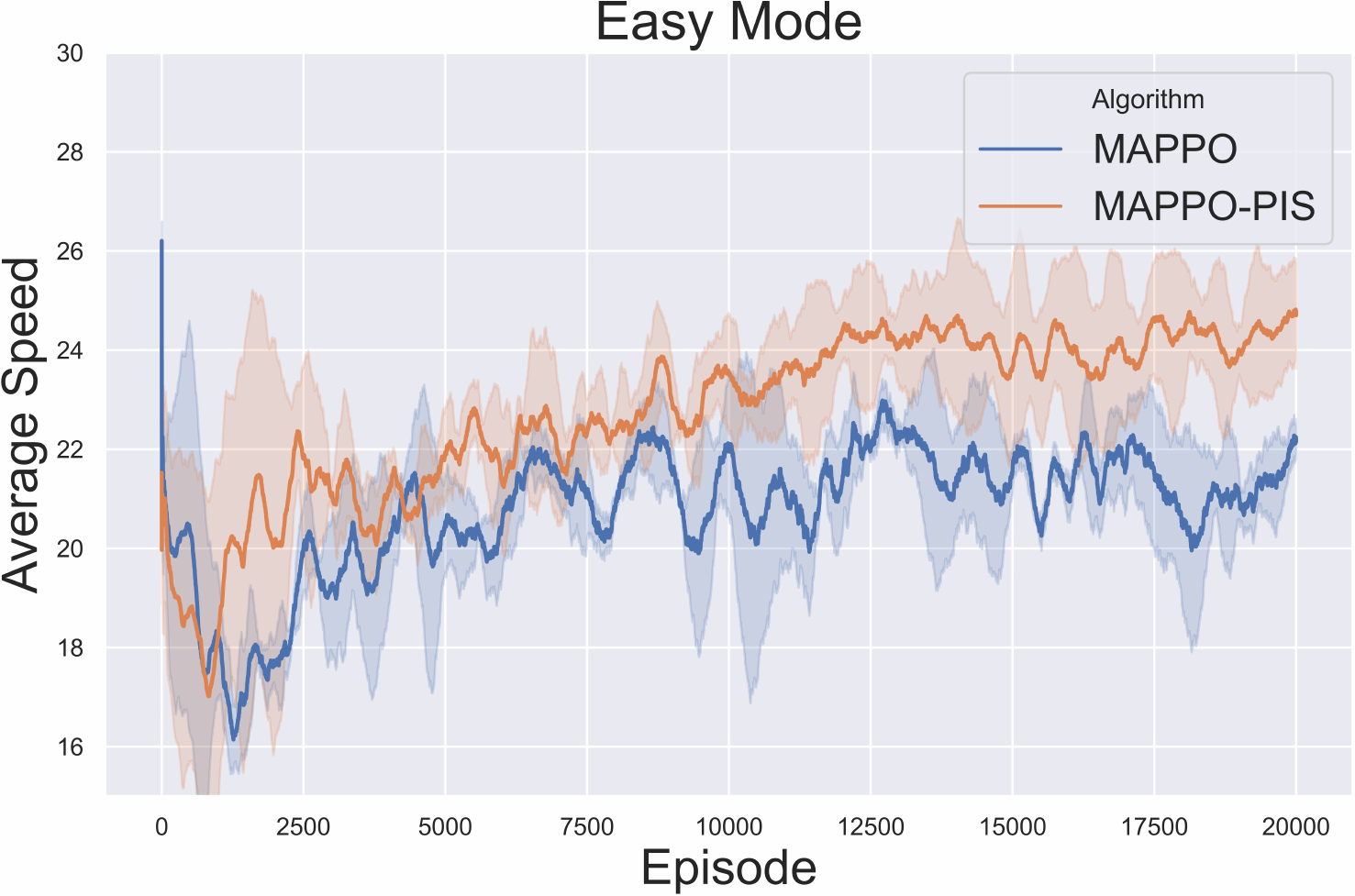} 
    \caption{}
    \label{fig:ab_speed_1}
  \end{subfigure}
  \begin{subfigure}{0.48\linewidth}
    \includegraphics [scale = 0.21]{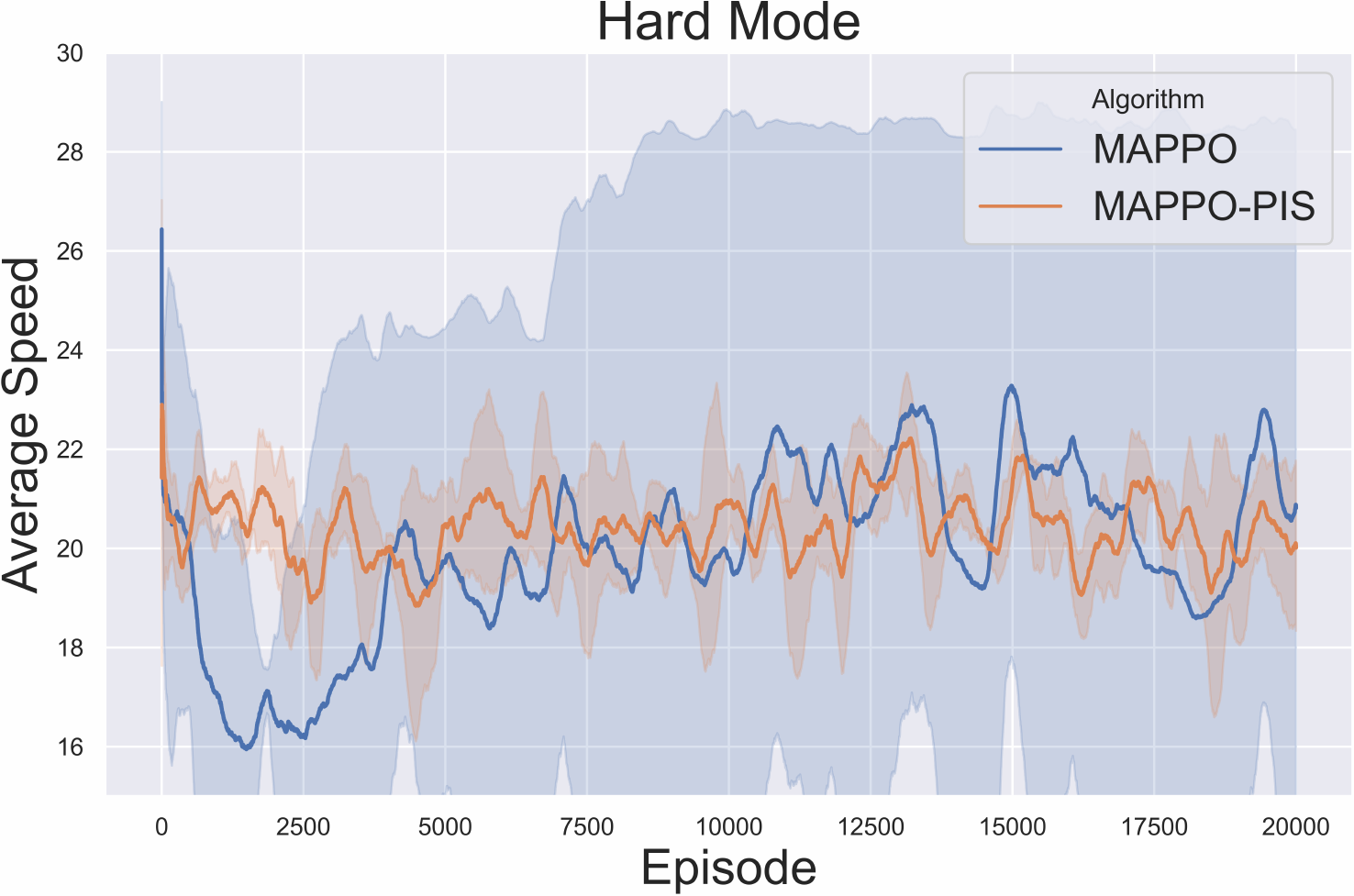}
    \caption{}
    \label{fig:ab_speed_2}
  \end{subfigure}
  \caption{Average speed under different traffic scenarios.}
  \label{fig:ablation_speed}
\end{figure}

{\bf Overall Performance}
\cref{fig:ablation_reward} demonstrates the average reward in the training process, under easy and hard traffic mode.
As shown in \cref{fig:ab_reward_1}, our proposed method improve the baseline slightly at first, however, average training reward of MAPPO-PIS grows quickly to around 65 after 10000th episode.
Similar result can be seen in \cref{fig:ab_reward_2}, our proposed method improves the baseline consistently during training process, which validates the effectiveness of our method.
Furthermore, our method also achieve efficient but stable performance, which can be seen in \cref{fig:ablation_speed}. 
In easy mode, the average CAVs' speed of our method is up to 24.5m/s, while the speed of MAPPO is 22.5m/s.
In hard mode, due to higher traffic density, average velocity of both methods is around 20m/s.
However, our method show more consistent performance as it has smaller standard deviation.
Overall, our proposed method exhibits higher efficiency in terms of the learning process and the vehicles in the scenario.

{\bf Safety Analysis.} \label{safety_analysis}
To prove the safety of our method, we conduct 30 random scenario tests on both algorithm under different traffic modes. 
In these experiments, collision rates are recorded and demonstrated in \Cref{tab:eva_base_and_ab}, which defined as the proportion of collision steps to the total number of steps.
As seen from \Cref{tab:eva_base_and_ab}, our model's collision rates under easy mode and hard mode are 0.00 and 0.01 respectively. 
While MAPPO also perform well in the easy mode with the collision rate 0.00, its collision rate increase to 0.03 in the hard mode.

Meanwhile, we also evaluate robustness of MAPPO-PIS, comparing to MAPPO.
\Cref{tab:hete_ab} show the performance of algorithms in different scenarios which contain heterogeneous vehicles.
Comparing to scenarios with homogeneous HVs, our method shows a slight increase in collision rate, however, its safety performance still exceeds that of MAPPO. 

Overall, our simulation results and analysis indicate that MAPPO-PIS can improve the safety performance in different modes of traffic, 
and our method is also robust to heterogeneous vehicles.


\begin{table}[tb]
  \caption{Testing performance comparison between the proposed method and 2 state-of-art baselines.
  }
  \label{tab:eva_base_and_ab}
  \centering
  \small
  % \resizebox{\textwidth}{!}{
  \setlength{\tabcolsep}{1.5mm}{
  \scalebox{0.90}{
  \begin{tabular}{@{}cccccc@{}}
    \toprule
    \textbf{Scenarios} & \textbf{Metrics} & \textbf{MAPPO-PIS} & \textbf{MAPPO} & \textbf{MAACKTR} & \textbf{MAA2C} \\ 
    \midrule
    \multirow{3}{*}{Easy Mode} & Evaluation Reward & \textbf{73.45} & 44.03 & -2.92 & 36.27 \\ 
    & Collision Rate & \textbf{0.00} & 0.00 & 0.01 & 0.01 \\ 
    & Average Speed (m/s) & \textbf{25.69} & 25.06 & 25.59 & 23.98 \\ 
    \midrule
    \multirow{3}{*}{Hard Mode} & Evaluation Reward & \textbf{33.27} & -36.58 & -17.12 & 0.30 \\ 
    & Collision Rate & \textbf{0.01} & 0.03 & 0.02 & 0.01 \\ 
    & Average Speed (m/s) & \textbf{21.61} & 23.63 & 18.65 & 13.75 \\ 
    \bottomrule
  \end{tabular}}}
  % }
\end{table}

\begin{table}[tb]
  \caption{Testing performance comparison between the proposed method and 2 state-of-art baselines in the scenarios with heterogeneous vehicles.}
  \label{tab:hete_ab}
  \centering
  \small
  \setlength{\tabcolsep}{7mm}{
  \scalebox{0.9}{
  \begin{tabular}{@{}cccccc@{}}
    \toprule
    \textbf{Scenarios} & \textbf{Metrics} & \textbf{MAPPO-PIS} & \textbf{MAPPO} \\ 
    \midrule
    \multirow{3}{*}{Easy Mode} & Evaluation Reward & \textbf{37.86} & 9.24 \\ 
    & Collision Rate & \textbf{0.01} & 0.02 \\ 
    & Average Speed (m/s) & \textbf{24.70} & 24.93 \\ 
    \midrule
    \multirow{3}{*}{Hard Mode} & Evaluation Reward & \textbf{8.17} & -48.24 \\ 
    & Collision Rate & \textbf{0.04} & 0.09 \\ 
    & Average Speed (m/s) & \textbf{21.26} & 23.75 \\ 
    \bottomrule
  \end{tabular}}}
\end{table}

\subsection{Algorithm Comparison}
In this subsection, two state-of-the-art MARL algorithms, including MAACKTR and MAA2C, are employed as baseline methods in our work, analyzing the effectiveness of MAPPO-PIS in different scenarios.

\begin{figure}[tb]
  \centering
  \begin{subfigure}{0.48\linewidth}
    \includegraphics[scale = 0.21]{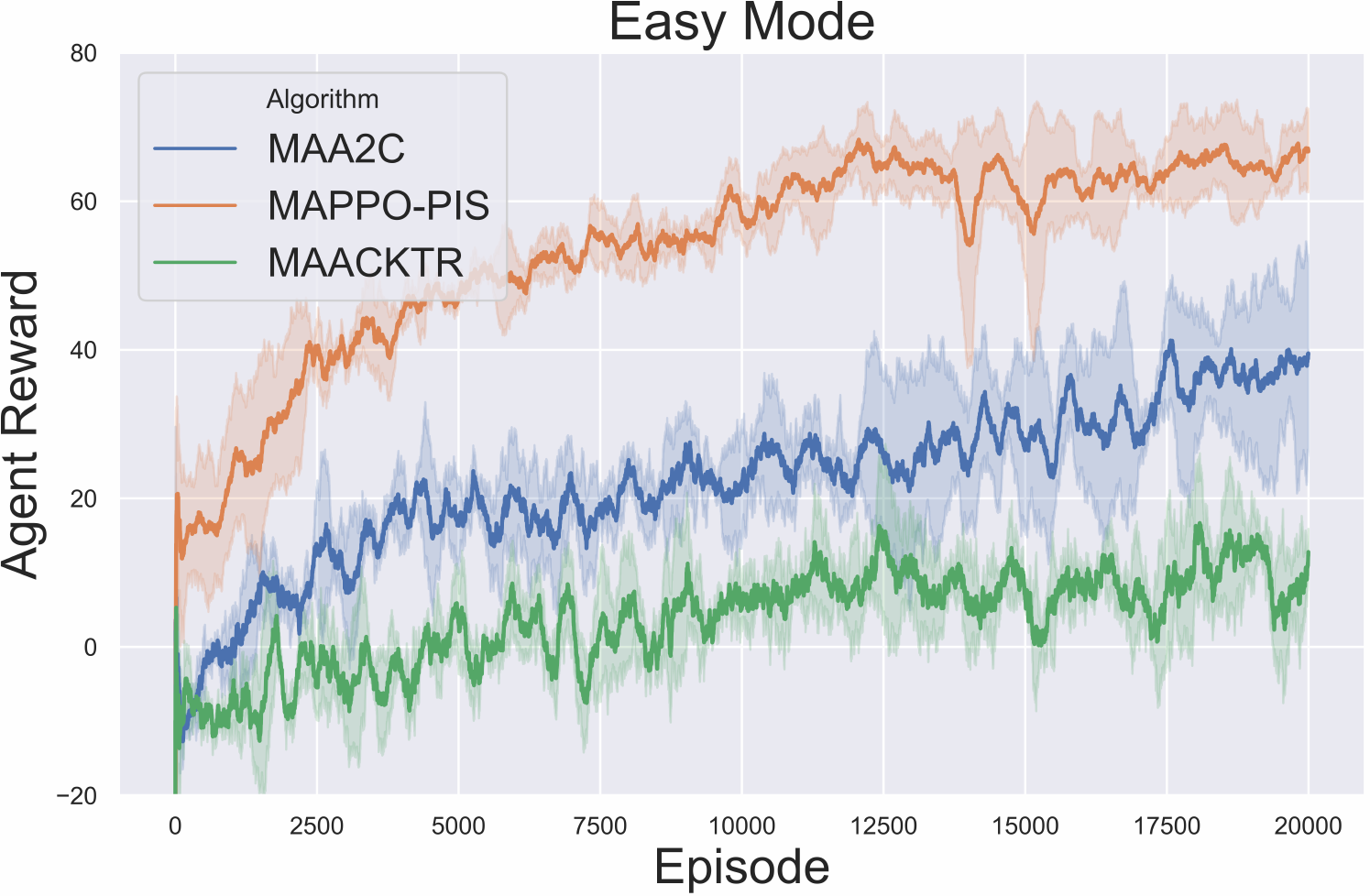}  
    \label{fig:base_reward_1}
  \end{subfigure}
  \hfill
  \begin{subfigure}{0.48\linewidth}
    \includegraphics[scale = 0.21]{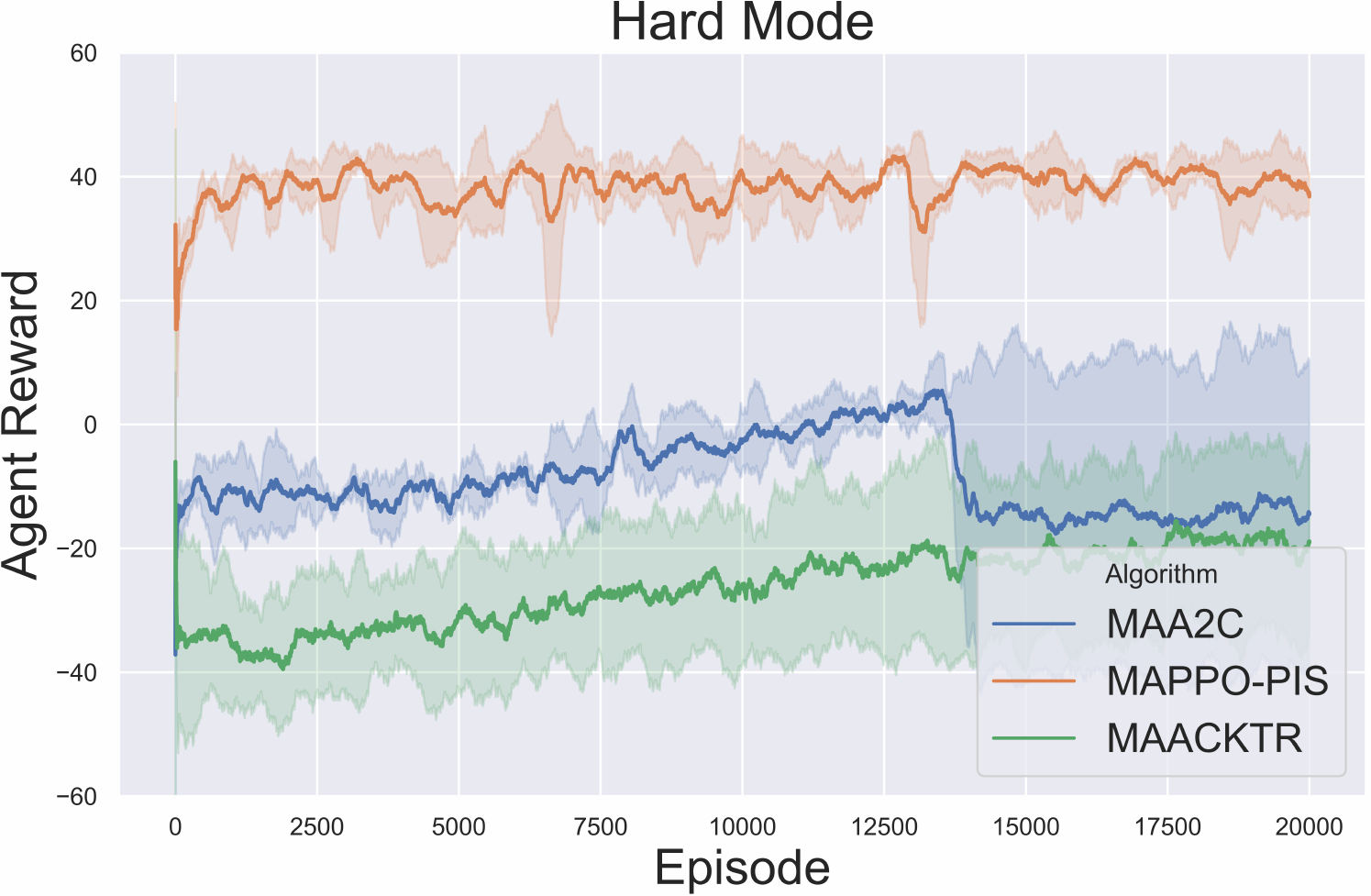}
    \label{fig:base_reward_2}
  \end{subfigure}
  \caption{Evaluation curves MAPPO-PIS and several state-of-art MARL baselines under different levels of traffic scenario.}
  \label{fig:base_reward}
\end{figure}

{\bf Overall Performance.}
As shown in \cref{fig:base_reward}, the proposed MAPPO-PIS significantly outperforms the baseline methods in terms of training average speed in both easy mode and hard mode.
By examining the safety of CAVs' intention and replacing those unsafe ones, the proposed method could avoid earlier terminations due to collisions, improving learning efficiency.

\begin{figure}[tb]
  \centering
  \begin{subfigure}{0.48\linewidth}
    \includegraphics[scale = 0.21]{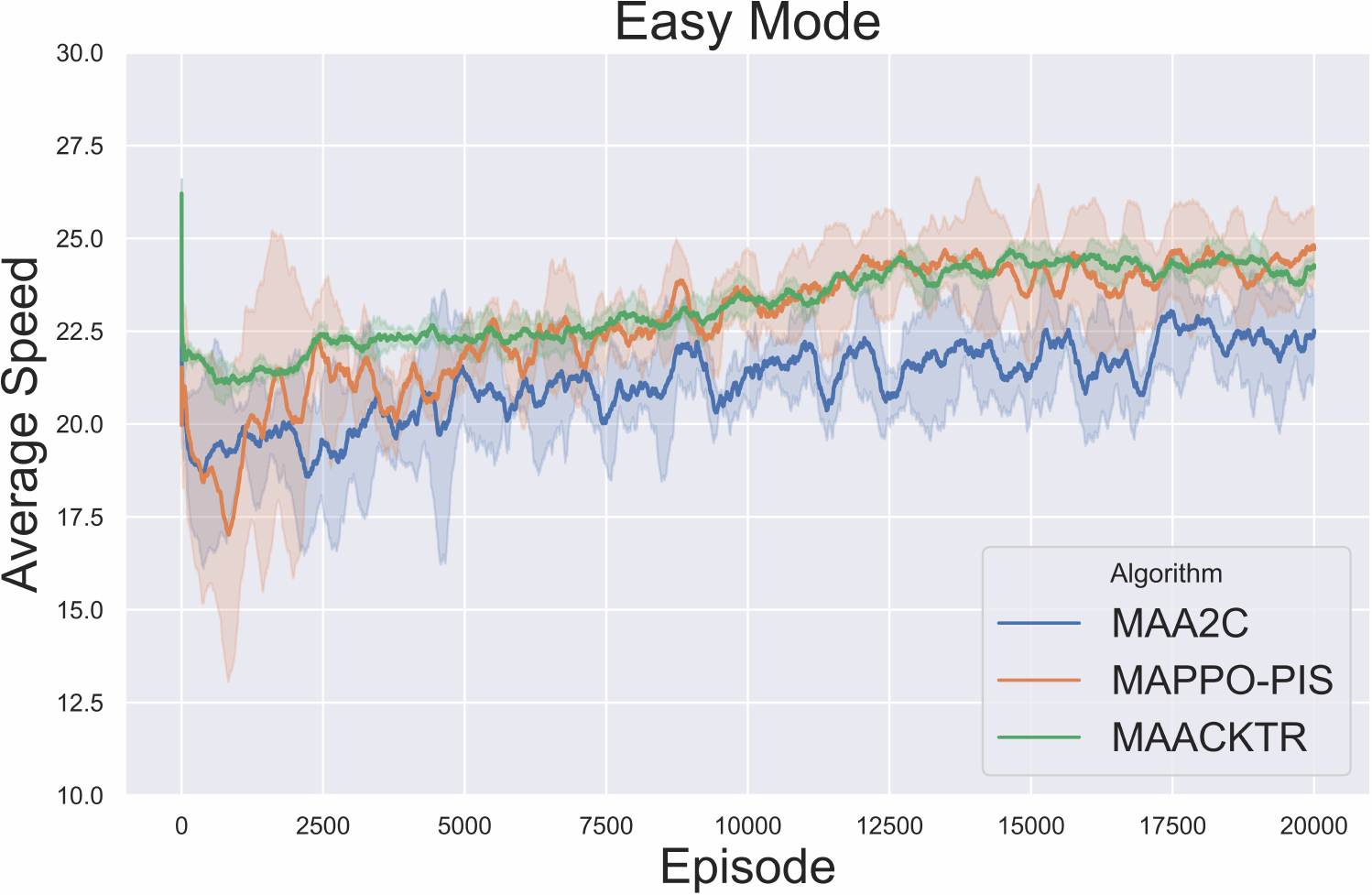}  
    \label{fig:base_speed_1}
  \end{subfigure}
  \hfill
  \begin{subfigure}{0.48\linewidth}
    \includegraphics[scale = 0.21]{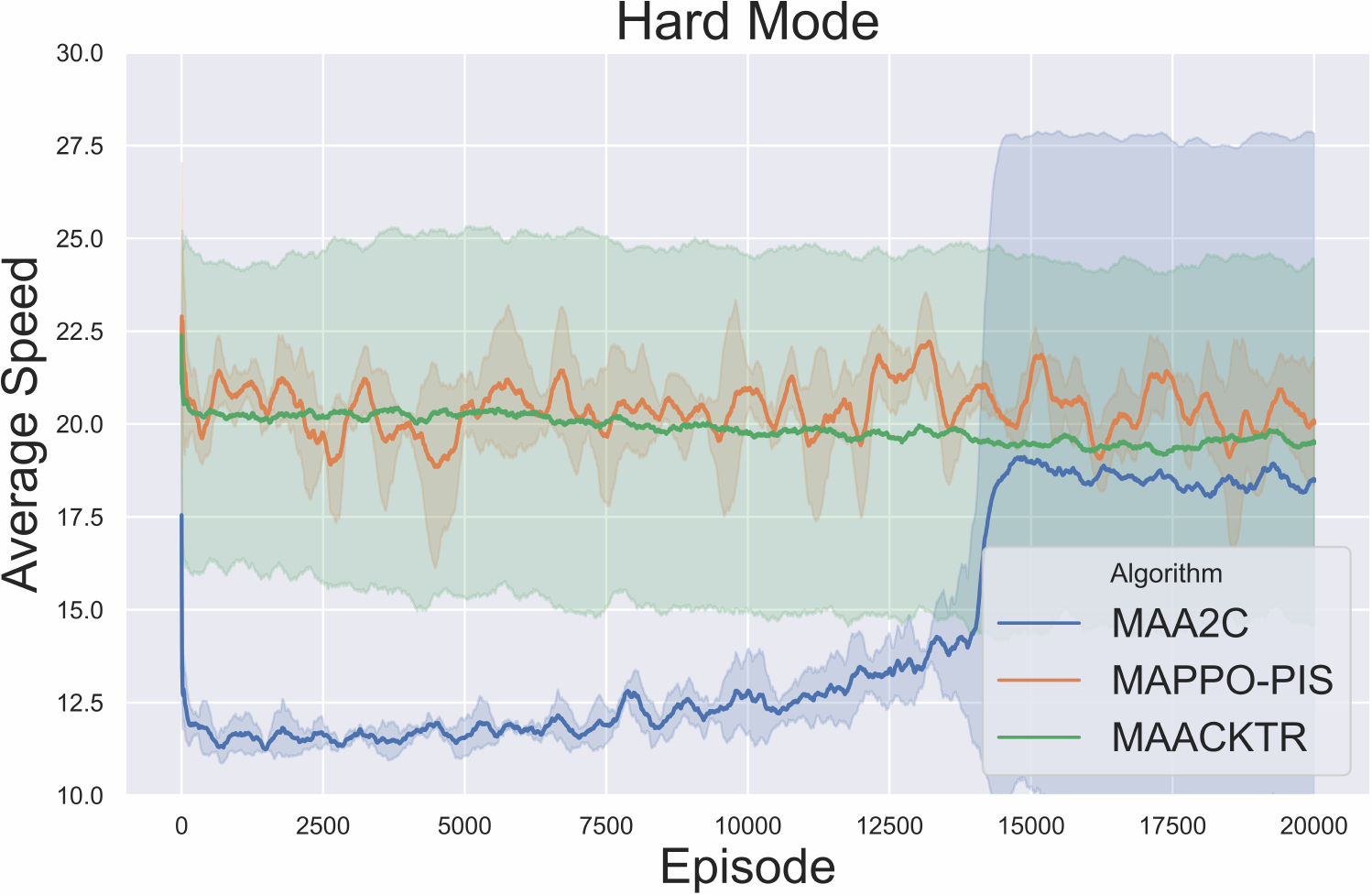}
    \label{fig:base_speed_2}
  \end{subfigure}
  \caption{Training average speed curves of our method and two state-of-art MARL algorithms}
  \label{fig:base_speed}
\end{figure}

The average speed of CAVs is depicted in \cref{fig:base_speed}.
It is clear that our method has similar average speed to MAA2C in both modes (\ie around 24m/s and 21m/s, respectively), higher than that of MAACKTR.
Furthermore, while our method shows a high speed in the hard mode, it also has smaller standard deviation, which mean that it has more stable performance.

To sum up, MAPPO-PIS show great performance in learning efficiency and traffic efficiency, and at the same time has strong stability.

{\bf Safety Analysis.}
After training, we evaluate the baseline model for 30 times, which is the same as the setting in \cref{safety_analysis}.
As shown in \cref{tab:eva_base_and_ab}, proposed method can run without collision in easy mode, while the collision rate is 0.01 for both baseline methods.
In the hard mode, due to the increase of traffic density, MAPPO-PIS see an increase in collision rate, which grow to 0.02, but it still outperforms the MAACKTR (which collision rate is 0.02).
It should be emphasized that, although the proposed method has the same collision rate as MAA2C, it outperforms the MARL baselines due to its higher average speed.
This indicates that IGM and SEM can enhance traffic efficiency while maintaining a lower collision rate.

\begin{figure}[tb]
  \centering
  \begin{subfigure}{0.48\linewidth}
    \includegraphics[scale = 0.21]{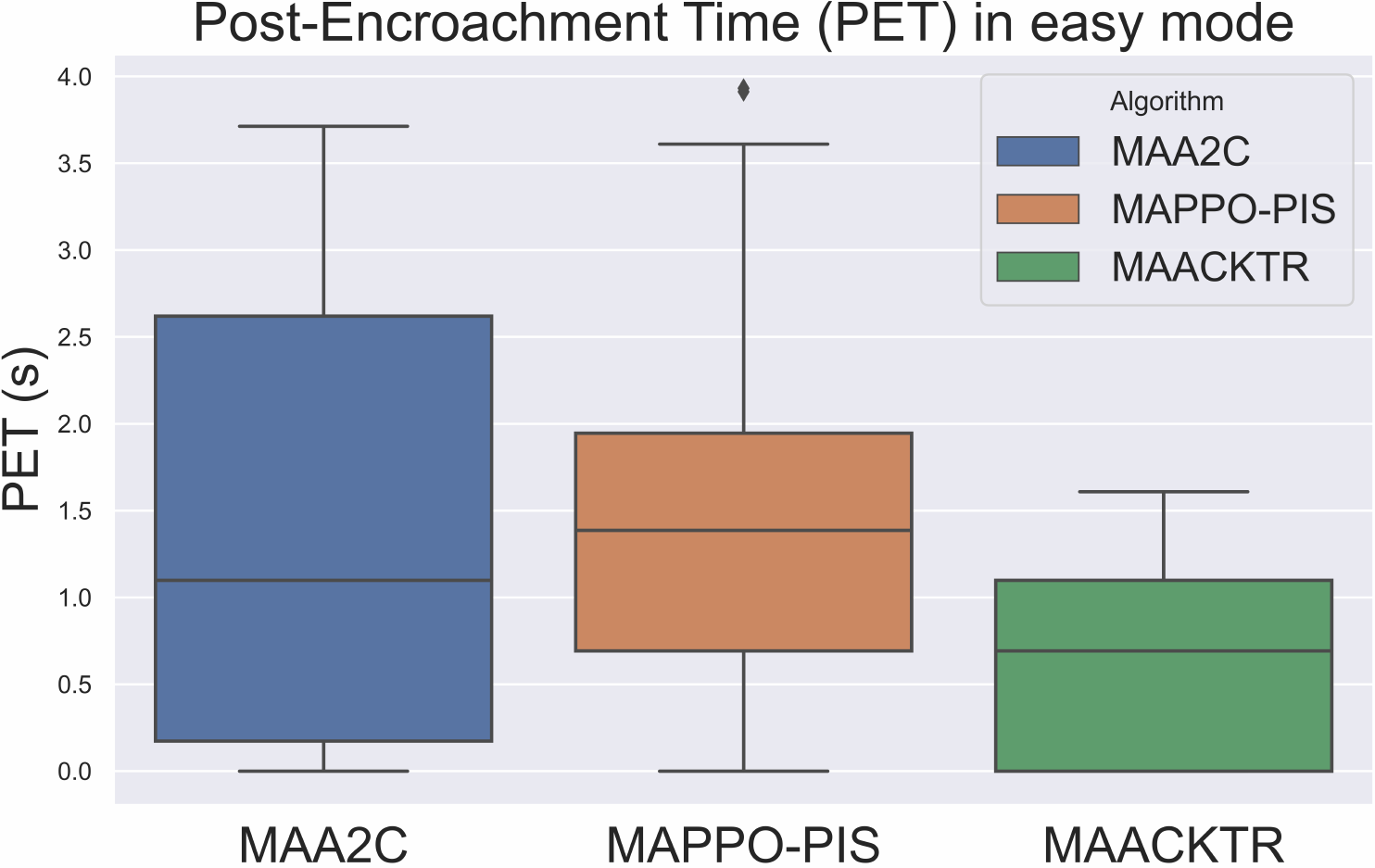} 
    \caption{}
    \label{fig:box_1}
  \end{subfigure}
  \hfill
  \begin{subfigure}{0.48\linewidth}
    \includegraphics [scale = 0.21]{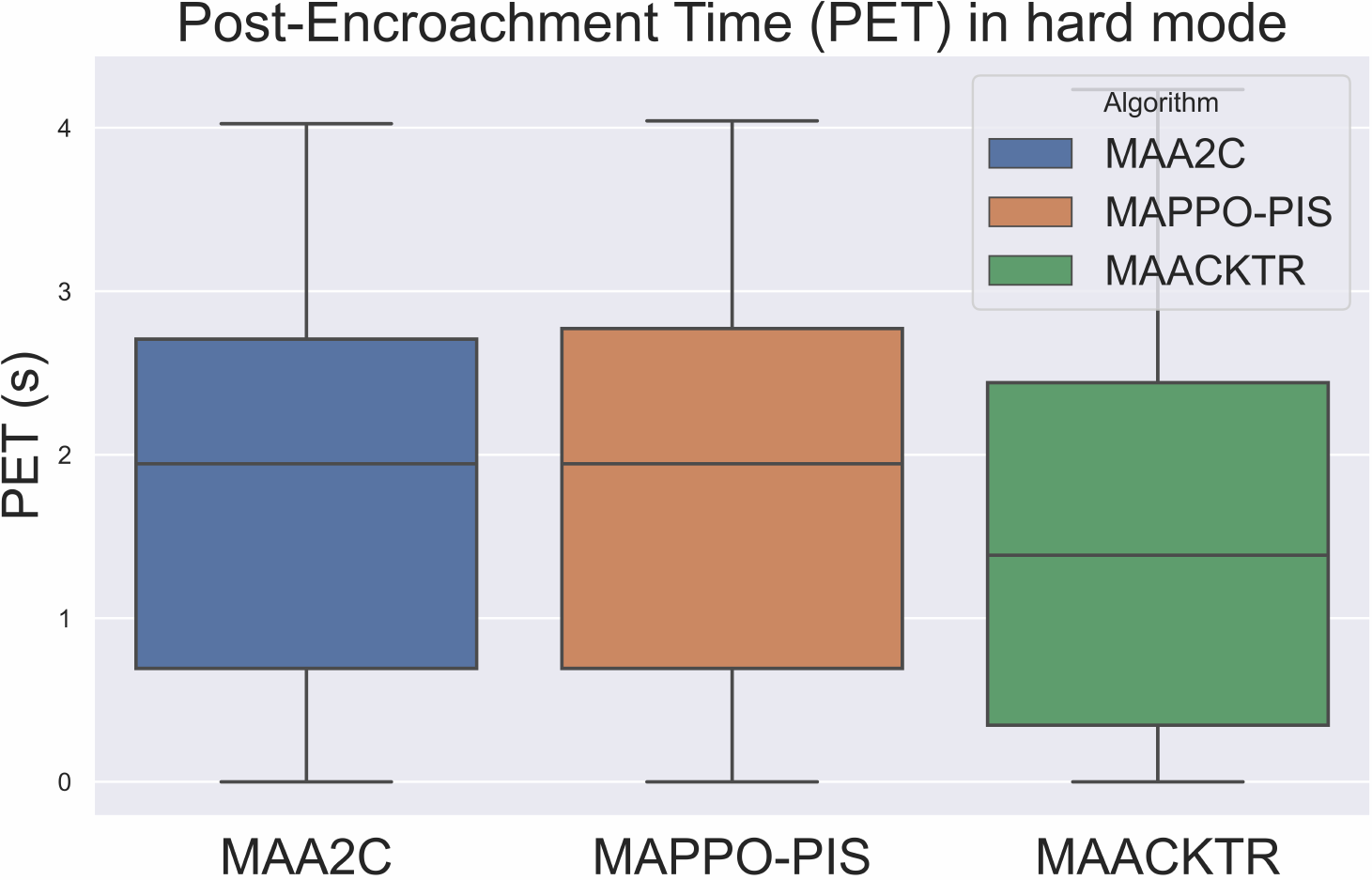}
    \caption{}
    \label{fig:box_2}
  \end{subfigure}
  \caption{CAVs’ PET values of different algorithms in different scenarios.}
  \label{fig:base_box}
\end{figure}

Meanwhile, We assess the safety of the merging area using the post-encroachment time (PET) metric\cite{42}.
\cref{fig:base_box} illustrates the PETs of different algorithms in the easy and hard traffic mode.
In easy mode, the average PET of our method, MAA2C and MAACKTR is 1.6s, 1.1s, and 0.65s respectively. It's obvious that our method could improve the safety performance in the easy scenario.
The average PET values of MAA2C and our method is 1.95s in the hard mode, higher than the average PET value of MAACKTR.

Overall, the simulation results and analysis demonstrate that our method surpasses several state-of-the-art MARL algorithms, achieving higher traffic efficiency and enhanced safety performance.

\subsection{Macro Analysis}
\begin{figure}[tb]
  \centering
  \begin{subfigure}{0.9\linewidth}
    \includegraphics[height = 2 cm]{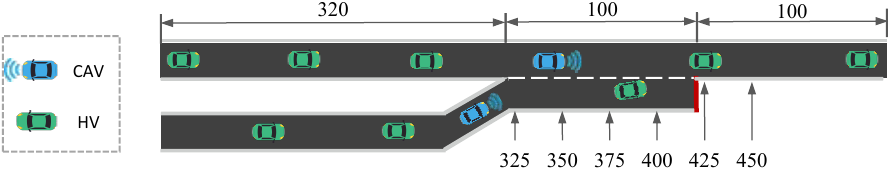}
    \caption{On-ramp merging scenario with coils}
    \label{fig:Coil}
  \end{subfigure}
  
  \vspace{0.25cm} % 调整此值以控制上下图之间的间距
  
  \begin{subfigure}{0.48\linewidth}
    \includegraphics[scale = 0.21]{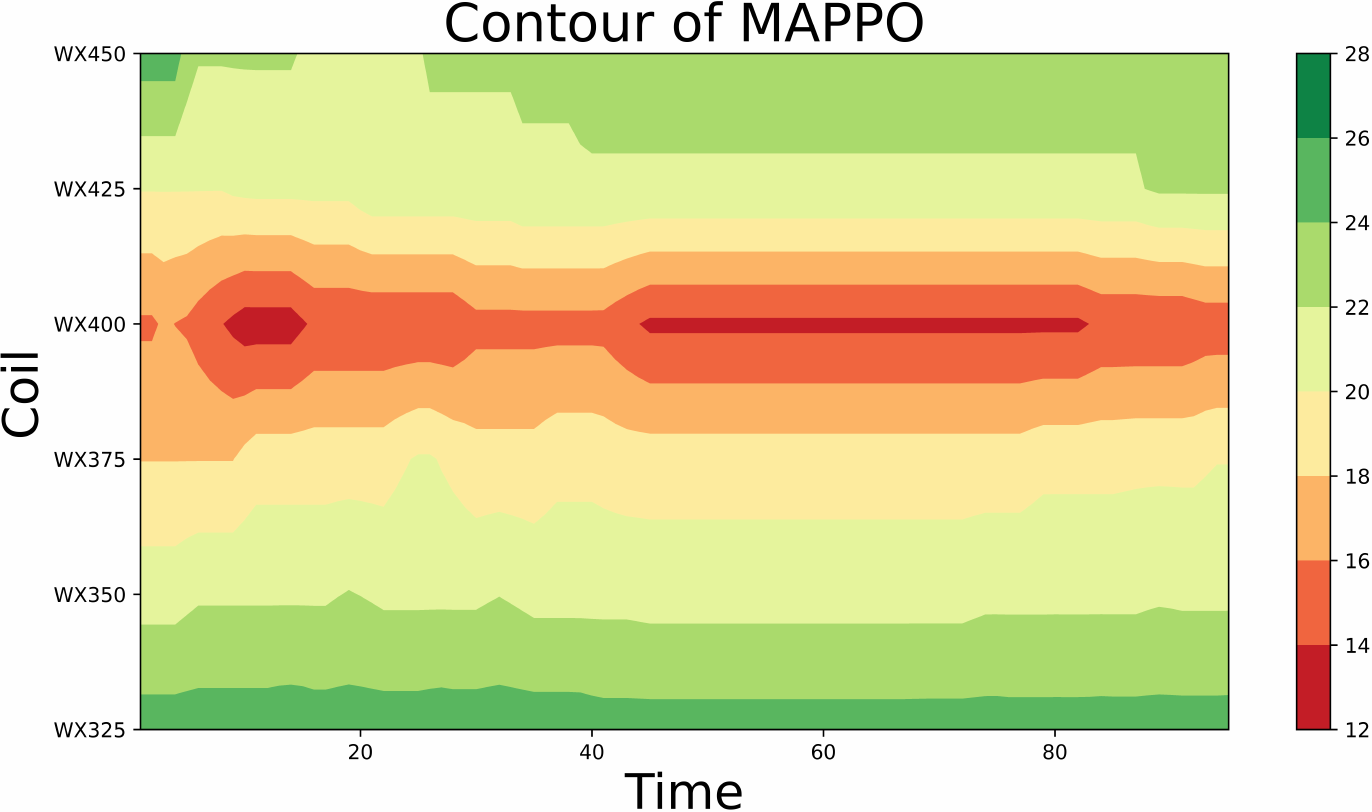}
    \caption{Contour map of MAPPO}
    \label{fig:contour_mappo}
  \end{subfigure}
  \begin{subfigure}{0.48\linewidth}
    \includegraphics[scale = 0.21]{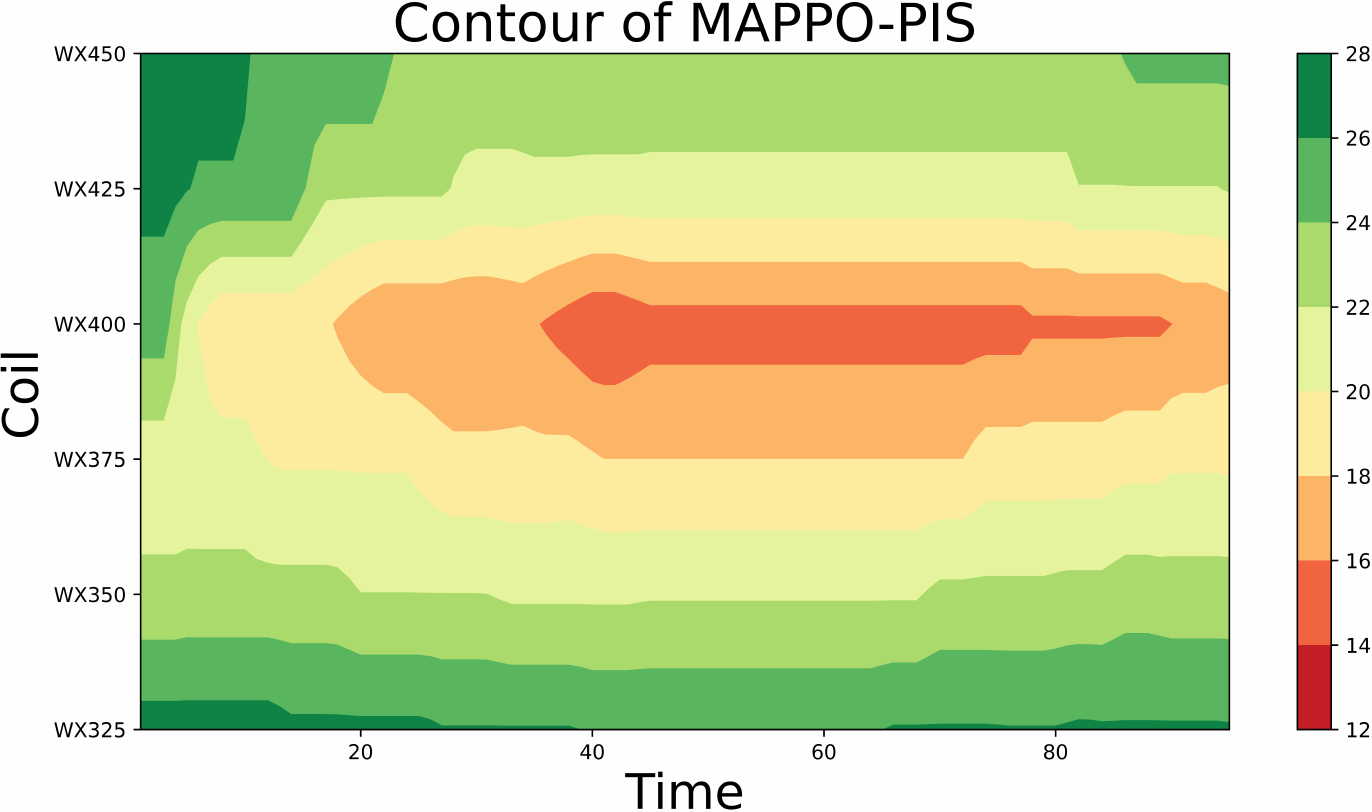}
    \caption{Contour map of  MAPPO-PIS}
    \label{fig:contour_our}
  \end{subfigure}
  
  \caption{Macro analysis of traffic flow in the merging area. }
  \label{fig:Macro Analysis}
\end{figure}

To visualize the effect of MAPPO-PIS on the traffic flow in the merging area, we set several virtual coils to record the average speed of vehicles on the road, which is illustrated in \cref{fig:Coil} (\ie 325m to 450m at 25m intervals). 
Average speeds over 100 time steps are recorded, ranging from 12 to 28m/s. 
In this work, when the time mean speed (\ie TMS, average speed of cross-section) falls below 16m/s, we consider the bottleneck to have failed.

\Cref{fig:Macro Analysis} shows the contour plots of time mean speeds in the merging area, which use MAPPO and MAPPO-PIS, respectively.
On the left side, it can be seen that the bottleneck breakdown occurs roughly from 5th time step in coil 400, and then become worse around 10 to 15th time-step and 45 to 85th time-step.
In contrast, our method begin to fail until 40th time-step, and it soon dissipated at about 85th time-step. 
At the same time, the TMS of our method is faster than that of MAPPO, which shows the effectiveness of our MAPPO-PIS.

\section{Conclusion}
In this paper, the MAPPO-PIS method is proposed for coopertative driving of CAVs in the human-machine mixed merging area, alleviating the effect of bottleneck breakdown.
The key components of MAPPO-PIS are Intention Generator Module (IGM) and Safety Enhanced Module (SEM).
In IGM, future intentions of CAVs are generated, and then input them to SEM, which could detect their safety margin and correct them into safe intention accordingly.
Experiments in highway-env simulator show that, compared to all other baselines, the proposed method achieve better performance in terms of learning efficiency, safety and robustness, highlighting its potential for enhancing traffic flow and safety in mixed driving environments.


\section*{Acknowledgements}
This work was supported in part by the State Key Lab of Intelligent Transportation System under Project No. 2024-A002, the Shanghai Scientific Innovation Foundation (No.23DZ1203400), the State Key Laboratory of Intelligent Green Vehicle and Mobility under Project No. KFZ2408, and the Fundamental Research Funds for the Central Universities.

\bibliographystyle{splncs04}
\bibliography{main}
\vfill
\newpage

\section*{Appendix}
\appendix
\section{Problem Formulation}\label{Problem Formulation}
% the \\ insures the section title is centered below the phrase: AppendixA

\textbf{State Space:} 
Owing to the limitation of sensor hardware, the CAV in the on-ramp merging scenario can only detect the state information of the vehicles around it. 
We denote the state space of agent $i$ as ${s}_{i}$, which is a matrix with dimensions of ${N}_{{\mathcal{N}}_{i}} \times W$. 
In the state matrix, ${N}_{{\mathcal{N}}_{i}}$ is the number of surrounding vehicles which in the perception range of agent $i$, and $W$ is the number of features which is used to represent the state of vehicles, including the following five features:
\begin{itemize}
\item $isobservable$: a binary identifier indicating whether a vehicle is observable in the perception range of agent $i$.
\item ${x}_{l}$: longitudinal position of the observed vehicle.
\item $y$: lateral position of the observed vehicle.
\item ${v}_{x}$: longitudinal speed of the observed vehicle.
\item ${v}_{y}$: lateral speed of the observed vehicle.
\end{itemize}

Here, we define the "Surrounding Vehicles" as the nearest ${N}_{{\mathcal{N}}_{i}}$  vehicles within the ego vehicle's perception range, which is within 150 $m$ of longitudinal distance from the ego vehicle. In the on-ramp merging scenario, we set ${N}_{{\mathcal{N}}_{i}}$  as the setting in\cite{4}.
The whole state space of system is the combination of each agent's state, \ie $\mathcal{S}=\mathcal{S}_{1}\times \mathcal{S}_{2}\times \cdots \times \mathcal{S}_{N}$.
    
\textbf{Action Space:} \label{Action}
We define the action space $\mathcal{A}_{i}$ for agent $i$ as the set of high-level control decisions, including {\textit{turn left, turn right, cruising, speed up, slow down}}. 
After selecting one high-level decision, the lower-level controllers produce the corresponding steering and throttle control signals controlling CAVs' action. 
The action space of the system is the joint actions of all CAVs, \ie $\mathcal{A}=\mathcal{A}_{1}\times \mathcal{A}_{2}\times \cdots \times \mathcal{A}_{N}$.

\textbf{Reward Function:}    
The reward function has a great effect on the performance of the algorithm and the RL agents' behaviors. In this paper, aiming to make all the agents pass the merging area safely and efficiently, the reward of $i$th agent at the time step $t$ is defined as follows:
\begin{equation}
  {r}_{i,t}={\omega }_{c}{r}_{c}+{\omega }_{s}{r}_{s}+{\omega }_{h}{r}_{h}+{\omega }_{m}{r}_{m}.
\end{equation}
where ${\omega }_{c},{\omega }_{s},{\omega }_{h}$ and ${\omega }_{m}$ are positive weighting coefficients of collision reward ${r}_{c}$, stable-speed reward ${r}_{s}$, headway cost reward ${r}_{h}$ and merging cost reward ${r}_{m}$, respectively.
These evaluation terms are defined as follows:
\begin{itemize}
\item Collision Reward ${r}_{c}$:
Safety is the most essential criterion for all the vehicles. To achieve safe driving, we define the collision reward with greater penalty when a collision occurs, which is defined as:
\begin{equation}
  {r}_{c}=\left\{\begin{matrix}
-1,& \text{If collision happened,}\\
 0,& \text{Otherwise.}
\end{matrix}\right.
\end{equation}
\item  Stable-Speed Reward ${r}_{s}$: 
Based on the speed recommendation from US Department of Transportation (\ie20-30m/s \cite{30}) and the speed range in the Next Generation Simulation (NGSIM) dataset (\ie the minimum speed is at 6-8m/s\cite{31}),
we set the minimum and the maximum speeds of the ego vehicles in the merging area as 10m/s and 30m/s respectively.
In order to evaluate the stability and efficiency of the performance of the ego vehicles, we defined the stable-speed reward as
\begin{equation}
  {r}_{s}=\min \begin{Bmatrix}
 \frac{{v}_{t}-{v}_{min}}{{v}_{max}-{v}_{min}},\ 1
\end{Bmatrix}.
\end{equation}
\item Headway Cost Reward ${r}_{h}$:
To ensure the safety of the current ego vehicle and the preceding vehicle, we set the headway cost reward as follow:
\begin{equation} \label{Eq:rh}
  {r}_{h}=\log_{}{\frac{{d}_{headway}}{{t}_{h}{v}_{t}}}. 
\end{equation}
where ${d}_{headway}$ is the distance headway and ${t}_{h}$ is the preset time headway threshold. 
Following to the time headway threshold suggested in\cite{32}, we choose ${t}_{h}$ as 1.2 \textit{s} in this paper.
According to the equation, the ego vehicle will be penalized when the time headway is less than ${t}_{h}$, and rewarded when it is greater than ${t}_{h}$.
\item Merging Cost Reward ${r}_{m}$: 
In order to avoid deadlocks in the merging area\cite{33}, we designed the merging cost reward, which can penalize the ego vehicle according to its waiting time on the merge lane. 
It is defined as:
\begin{equation}
  {r}_{m}=\exp_{}{}{\frac{-{(x-L)}^{2}}{10L}}.
\end{equation}
\end{itemize}
where we set $x$ is the distance that the ego vehicle has traveled on the ramp and $L$ is the total length of the ramp, which can be seen in \cref{fig:ramp}.
\vfill
\newpage
\section{PSEUDO CODE} \label{Re: PSEUDO CODE}

\begin{algorithm}[!h]
    \caption{Safety Enhanced Module (SEM)}
    \label{alg:SEM algorithm}
    \begin{algorithmic}[1]
    \renewcommand{\algorithmicrequire}{\textbf{Parameter:}}
    \renewcommand{\algorithmicensure}{\textbf{Output:}}
        \REQUIRE{${\alpha }_{1}$, ${\alpha }_{2}$, ${\alpha }_{3}$, ${T}_{n}$, $L$, ${d}_{headway}$, ${t}_{h}$}
        \ENSURE{Safe intention ${a}_{t}'$}
        \STATE Initialize the priority list ${P}_{t}$
        \FOR{$i=0$ to $N-1$}
        \IF{CAV $i$ on the ramp}
        \STATE Compute merging metric ${p}_{m}$, merging-end metric ${p}_{e}$ and headway metric ${p}_{h}$, according to \cref{Eq:pm}, \cref{Eq:pe} and \cref{Eq:ph}, respectively
        \ELSE
        \STATE Compute headway metric ${p}_{h}$, according to \cref{Eq:ph}
        \ENDIF
        \STATE Compute  the priority score of CAV $i$ according to \cref{eq:priority}
        \STATE Add CAV $i$ and its priority score to priority list ${P}_{t}$
        \ENDFOR
        \STATE Sort the priority list ${P}_{t}$ according to their priority scores in descending order
        \FOR{$j=0$ to $\left | {P}_{t}\right |-1$}
        \STATE Obtain ${P}_{t}[j]$
        \STATE Obtain its surrounding vehicles ${\mathcal{N}}_{{P}_{t}[j]}$ within the perception range
        \STATE Predict surrounding vehicles' intention trajectories ${\tau }_{v}$ for ${T}_{n}$ time steps.
        \IF{${\tau }_{v}$ and CAV $j$'s intention trajectory ${\tau }_{{P}_{t}[j]}$ collide}
        \STATE Replace the unsafe intention to Safe intention ${a}_{t}^{'}$ according to \cref{Eq: action replace}
        \STATE Regenerate CAV $j$'s intention trajectory ${\tau }_{{P}_{t}[j]}$
        \ENDIF
        \ENDFOR     
    \end{algorithmic}
\end{algorithm}
\end{document}